\documentclass{article}
\usepackage{fullpage}
\usepackage[T1]{fontenc}
\usepackage{microtype}
\usepackage[misc]{ifsym}
\usepackage{graphicx}
\usepackage{mathtools}
\usepackage{xcolor}
\usepackage{xparse}
\usepackage{float}
\usepackage{siunitx}
\usepackage{booktabs}
\usepackage{xfp}
\usepackage{threeparttable}
\definecolor{LightBlue}{HTML}{00F9DE}
\usepackage{wrapfig}
\usepackage{authblk}

\usepackage{algorithm}
\usepackage{algpseudocode}
\algtext*{EndFor}%
\algtext*{EndWhile}%
\algtext*{EndIf}%

\usepackage[caption=false]{subfig}
\usepackage{hyperref}

\newcommand{\Rreordered}{\pi_R(R_i)}
\newcommand{\Creordered}{\pi_C(C_i)}

\newcommand{\cons}{\mathit{cons}}
\newcommand{\consR}{\cons(\Rreordered)}
\newcommand{\consC}{\cons(\Creordered)}

\newcommand{\nonzero}{\mathit{nonzero}}

\DeclarePairedDelimiter{\abs}{\lvert}{\rvert}

\newcommand{\clusters}{\mathit{clusters}}
\newcommand{\demerit}{\mathit{demerit}}

\newcommand{\similarity}{\mathit{similarity}}
\newcommand{\density}{\mathit{density}}
\newcommand{\rowBlock}{\mathit{block}_R}
\newcommand{\colBlock}{\mathit{block}_C}

\newcommand{\proximity}{\operatorname{prox}}
\newcommand{\consecutiveClusterArea}{\operatorname{clArea}}
\newcommand{\clusterArea}{\consecutiveClusterArea}
\newcommand{\uninterruptedArea}{\operatorname{uninter}}
\newcommand{\uninterrupted}{\uninterruptedArea}

\newcommand{\averageRandomScore}{\mathit{averageRandomScore}}

\begin{document}
\title{Visualizing Overlapping Biclusterings and Boolean Matrix Factorizations}%
\date{}
\author[1]{Thibault Marette}
\author[2]{Pauli Miettinen}
\author[1]{Stefan Neumann}
\affil[1]{KTH Royal Institute of Technology, Stockholm, Sweden \\ \{marette,neum\}@kth.se}
\affil[2]{University of Eastern Finland, Kuopio, Finland \\ pauli.miettinen@uef.fi}
\maketitle              %
\begin{abstract}
Finding (bi-)clusters in bipartite graphs is a popular data analysis approach.
Analysts typically want to visualize the clusters, which is simple  as long as
the clusters are disjoint.
However, many modern algorithms find overlapping clusters, making visualization  more complicated.
In this paper, we study the problem of visualizing \emph{a given clustering} of overlapping clusters in bipartite graphs and the related problem of visualizing Boolean Matrix Factorizations.
We conceptualize three different objectives that any good visualization should satisfy:
(1) proximity of cluster elements, (2) large consecutive areas of elements from the same cluster,
and (3) large uninterrupted areas in the visualization, regardless of the cluster
membership.
We provide objective functions that capture these goals  and algorithms that optimize these objective functions.
Interestingly, in experiments on real-world datasets, we find that the best trade-off between these competing goals is achieved by a novel heuristic, which locally aims to place rows and columns with similar cluster membership next to each other.

\end{abstract}
\section{Introduction}

Finding biclusters in bipartite graphs has been
studied for several decades~\cite{hartigan72direct,zha01bipartite} and it is
closely related to other problems, such as
co-clustering~\cite{dhillon01coclustering} and Boolean Matrix
Factorization~\cite{miettinen08discrete}.  While the goal of classic methods is to find
mutually disjoint biclusters, i.e., each vertex appears in at most one
bicluster, modern methods allow for \emph{overlap}: vertices can appear in
multiple clusters
\cite{hess2021broccoli,neumann18bipartite,miettinen08discrete,miettinen2020recent,1324618}.

To assess the outputs of biclustering algorithms, it can 
be helpful to visualize their outputs.
If all clusters are 
\emph{disjoint}, one  can  plot the biclusters one after
another in an arbitrary order. If clusters \emph{overlap}, the
visualization task becomes more difficult~\cite{Vehlow2017VisualizingGS}: it might not be possible to draw all biclusters as consecutive rectangles, as is the case in Fig.~\ref{fig:csqblock}, forcing the visualization to choose which clusters to split up.

\begin{figure}[t]
\centering
\subfloat[Dialect dataset, $k=54$\label{dialectPCV48}]{%
\begin{minipage}{.65\textwidth}
\centering
\includegraphics[height=.75\textwidth,angle=90]{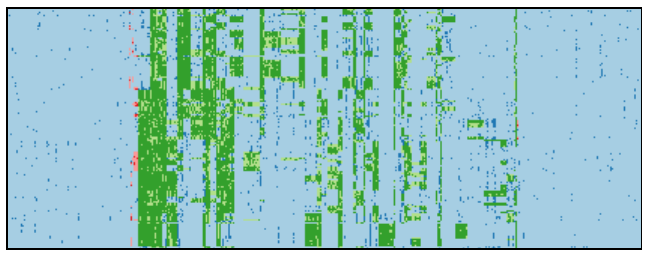}
\includegraphics[height=.75\textwidth,angle=90]{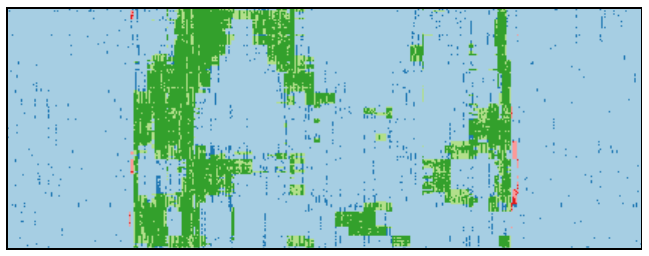}
\end{minipage}
}\hfil
\subfloat[Paleo dataset, $k=50$\label{paleo50}]{
\begin{minipage}{.29\textwidth}
\centering
\includegraphics[width=.75\textwidth]{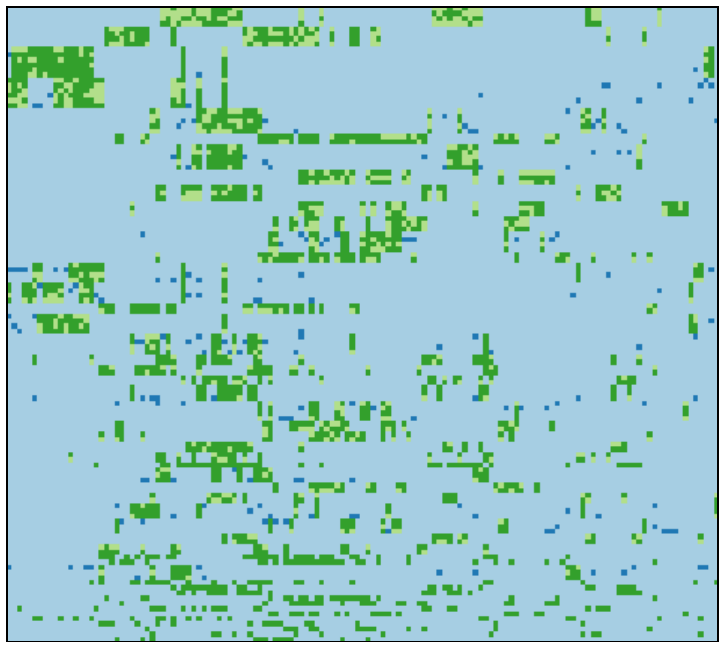}
\includegraphics[width=.75\textwidth]{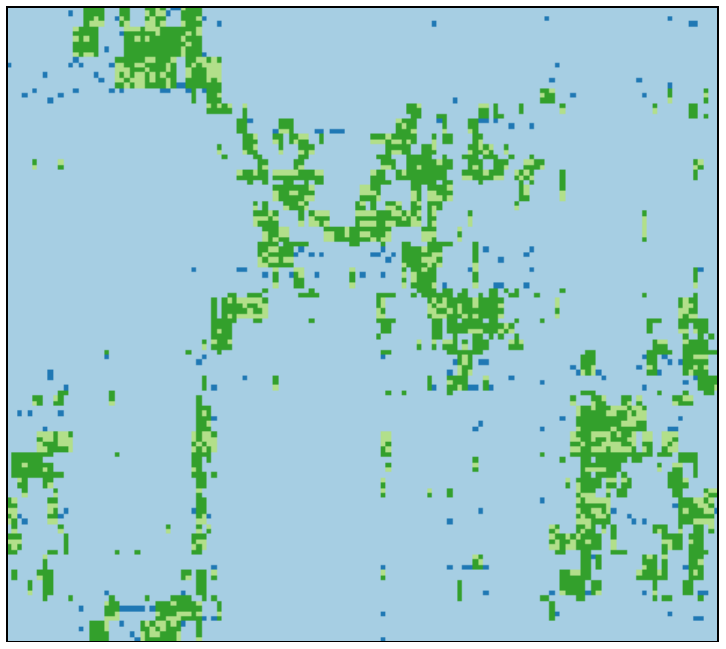}
\end{minipage}
}
\caption{Visualization of the same biclustering using ADVISER~\cite{colantonio2011visual} (top) and
	our TSP-based heuristic (bottom). The pictures at the bottom contain larger uninterrupted areas,
	which makes it easier to assess the structure in the data.%
 }
\label{fig:beat-adviser}
\end{figure}

This problem was studied in earlier work~\cite{jin2008overlapping,colantonio2011visual}, with the main goal of optimizing the proximity of elements that
belong to the same bicluster. However, this notion has drawbacks as biclusters
which are similar in one dimension but non-overlapping in another are
not incentivized to be visualized close to another. This leads to 
suboptimal visualizations for some biclusterings, as shown in
Fig.~\ref{fig:beat-adviser}.

In this paper, we revisit the problem of visualizing given
biclusterings. Rather than just looking at the proximity of elements from the
same bicluster, we identify three different aspects of good visualizations:
(1)~Proximity of elements from the same bicluster. (2)~Large consecutive areas
of elements from the same bicluster. (3)~Large uninterrupted areas in the
visualization, regardless of the bicluster membership. For each of these three
different aspects, we provide novel objective functions that allow us to
formally capture these intuitions. Especially Aspect~(3) will help us to bypass
the limitations from the approaches in~\cite{jin2008overlapping,colantonio2011visual}.

We also present several algorithms to optimize our objective functions.
As optimizing them directly is expensive in terms of time and
difficult in terms of quality, we present a novel heuristic which is
based on the concept of \emph{demerit}, which penalizes visualizations that
place rows and columns close to each other when they belong to different
biclusters.  We present experiments on real-world datasets which show that this
heuristic can be computed efficiently and that it provides a very good tradeoff
between the three objective functions, outperforming the method
from~Colantonio et al.~\cite{colantonio2011visual}.  In our experiments we focus
on medium-sized datasets, since visualizing large bipartite graphs requires
different methods~\cite{pezzotti18multiscale}.

Additionally, we introduce a novel post-processing step, which automatically
finds unclustered rows and columns that have high similarity with the provided biclusters. We believe that this will enable domain experts to
efficiently find structures that might have been missed by the original
biclustering algorithm.

We make our code\footnote{\url{https://github.com/tmarette/biclusterVisualization}\label{fn:code}}
and plots\footnote{\url{https://github.com/tmarette/VisualizingOverlappingBiclusteringsAndBMF-plots}\label{fn:plots}}
for all datasets available on GitHub. We note that, even though previous
works studied the question of visualizing overlapping biclusterings, none of
these works has its code available online.

\textbf{Related Work.}
Computing biclusterings of bipartite graphs is a classic problem that has been
studied at least since the 1970s~\cite{hartigan72direct} and it is related to
several other problems, such as bipartite graph
partitioning~\cite{zha01bipartite}, hypergraph
partitioning~\cite{alistarh15streaming}, bipartite stochastic block
models~\cite{neumann18bipartite} and co-clustering~\cite{dhillon01coclustering}.
It is also known that Boolean Matrix Factorization, which has been a popular
problem in the data mining
community~\cite{miettinen08discrete,lucchese10mining,hess17primping}, is closely
related~\cite{miettinen2020recent}.

Colantonio et al.~\cite{colantonio2011visual} studied the visualization of a given set of
overlapping biclusters as a biadjacency matrix. They introduced an
objective function, which optimizes the proximity of the rows and columns that
are contained in biclusters and which simultaneously tries to minimize gaps in
the visualization of each bicluster. They also proposed a greedy heuristic called
ADVISER for optimizing this objective function.  They experimentally showed
that their approach is superior to the approach by
Jin et al.~\cite{jin2008overlapping}, which only considers the perimeter of the
visualized biclusters.  The main drawback of the approach
in~\cite{colantonio2011visual} is that biclusters which are highly similar in
one dimension but are non-overlapping in another (e.g., they have overlapping
column clusters but non-overlapping row clusters) are not incentivized to be visualized close to
another.

Classic seriation methods~\cite{MatrixReorderingSTAR,Xu2016InteractiveVC} that
visualize biadjacency matrices are related to our work, but they do not support
visualizing a given input biclustering.  Leaf-ordering methods that visualize
dendrograms, e.g.,~\cite{Dendo}, can visualize a given hierarchical clustering,
but biclustering algorithms do not report a hierarchy of the biclusters
and thus these methods are not applicable. The BiVoC~algorithm~\cite{grothaus2006automatic} is also related, but its visualization repeats
rows and columns, which we do not permit here because visualizations with many repetitions quickly
become unclear.

We use biadjacency matrices to visualize biclusterings.
Alternatives include edge
bundlings~\cite{sun19effect,tatti19boolean} or anchored-maps~\cite{misue06drawing}.  Our
algorithms are completely unsupervised, but semi-supervised methods~\cite{xu16interactive} exist.

\section{Preliminaries}

Let $G = (R \cup C,E)$ be an unweighted, undirected bipartite graph and set
$m=\abs{R}$ and $n=\abs{C}$. We assume that $R=[m]$ and $C=[n]$,
where $[k]:=\{1,\dots,k\}$.  A \emph{biclustering} $((R_1,C_1),\dots,(R_k,C_k))$
of $G$ is a set of \emph{biclusters} $(R_i,C_i)$, where $R_i\subseteq R$ and $C_i\subseteq
C$ for all $i$. 
Note that this is a very general definition of biclustering: we do not
assume that the clusters~$R_i$ are mutually disjoint or that $\bigcup_i R_i = R$, and neither do we make these assumptions for the~$C_i$. Two biclusters $(R_i,C_i)$ and $(R_j,C_j)$ \emph{overlap} if
$R_i \cap R_j \neq \emptyset$ and $C_i\cap C_j\neq \emptyset$.

\textbf{Visualization.}
We visualize~$G$ using its $m \times n$ biadjacency matrix
$A\in\{0,1\}^{m\times n}$.  Note that the
vertices in $R$ correspond to the rows of $A$ and the vertices in $C$ correspond
to the columns of~$A$.  Thus, we will often refer to the clusters $R_i$ as the
\emph{row clusters} and to the clusters $C_i$ as the \emph{column clusters}.
When plotting~$A$, we use bright tiles for $1$-entries and dark tiles for
$0$-entries.

To visualize $A$, our goal is to find permutations $\pi_R \colon [m] \to [m]$
and $\pi_C \colon [n] \to [n]$ of the rows and columns of the biadjacency
matrix, respectively. Each element $r\in R$ ($c\in C$) is visualized in the
$\pi_R(r)$'th row ($\pi_C(c)$'th column) of the biadjacency matrix, i.e., we set
$A_{\pi(r),\pi(c)} = 1$ iff $(r,c)\in E$.

Throughout the paper we study the following problem. Given a bipartite graph
$G=(R\cup C, E)$ and a biclustering $((R_1,C_1),\dots,(R_k,C_k))$, find
permutations $\pi_R \colon [m] \to [m]$ and $\pi_C \colon [n] \to [n]$ of the
rows and columns that optimize an objective function, which encodes how well
the biclustering is visualized.

\textbf{Notation.}
Let $X$ be a set of integers and $\pi$ a permutation.  We write
$\pi(X)=\{\pi(x) : x\in X\}$ to denote $X$ under the permutation~$\pi$. We
write $\cons(X)$ to denote the partition of $X$ into maximal disjoint sets of
consecutive integers.  For instance, if $X=\{1,2,5\}$ then
$\cons(X)=\{\{1,2\},\{5\}\}$.  Note that if $R_i$ is a set of rows and $\pi_R$
is the row permutation, then $\pi_R(R_i)$ is the set of rows in which the
elements of $R_i$ are visualized; the sets of consecutive rows
(columns) in which elements from $R_i$ ($C_i$) are visualized is given by
$\consR$ ($\consC$).

Finally, for our algorithms it will be convenient to operate on \emph{row and
column blocks}. For brevity, we only give the definition for row blocks. The row blocks partition the sets of
rows, and they are defined such that each cluster can be expressed as the union
of a set of blocks. More formally, for $r\in[m]$ we let $\clusters_R(r) =
\{i : r\in R_i\}$ denote the set of indices of all row clusters that
contain row $r$. Now, the \emph{row block of $r$}
is given by $\rowBlock(r) = \{ r' : \clusters_R(r)=\clusters_R(r') \}$, i.e., it
is the set of all rows~$r'$
that are contained in exactly the same row clusters as $r$. Next, the set of row
blocks is given by $\mathcal{B}^R = \{\rowBlock(r) : r\in R\}$; see Fig.~\ref{fig:definitions} for an example. Given a row
block $b\in\mathcal{B}^R$ it will be convenient for us to write $\clusters_R(b)$
to denote the row clusters in which $b$~is contained, i.e., $\clusters_R(b) =
\clusters_R(r)$ for all $r\in b$. For column blocks, we define $\clusters_C(c)$,
$\colBlock(c)$ and $\mathcal{B}^C$ in the same way.

\textbf{Visualizing weighted and directed graphs.} The algorithm we propose in
this paper is tailored to visualize unweighted bipartite graphs, through their
Boolean biadjacency matrix~$A$.  We note that since in general~$A$ is
asymmetric, our algorithms can also be used to visualize the adjacency matrix of
directed graphs (note that in this case the set of row and column clusters will
be the identical).  It is also possible to use our algorithm to visualize
weighted graphs; however, in this case one has to make adjustments to the
coloring scheme to visualize the different weights (here, we focus on the
Boolean case in which we never need more than six colors).

\section{Visualization Objectives}
\label{sec:objectives}

In this section, we introduce our objective functions that measure different
aspects of how well a biclustering is visualized.

\begin{figure}[tbp]
\centering
\subfloat[Proximity\label{fig:prox}]{%
\begin{minipage}[t]{.3\textwidth}
  \centering
  \includegraphics[width=.45\textwidth]{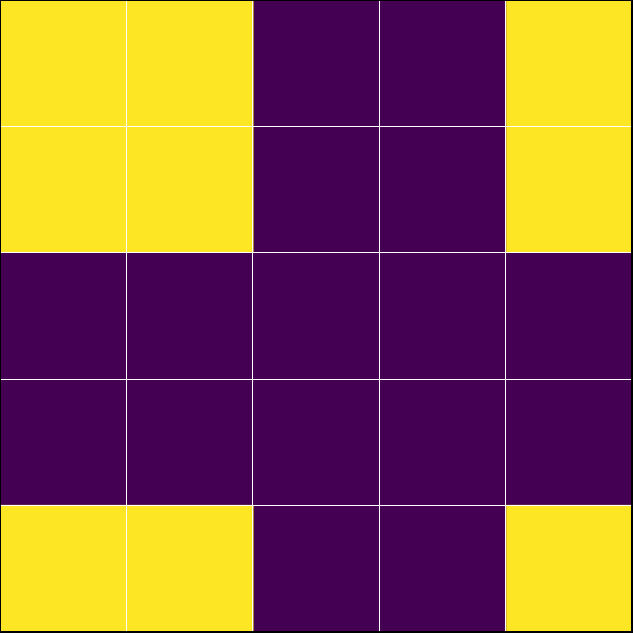}
  \includegraphics[width=.45\textwidth]{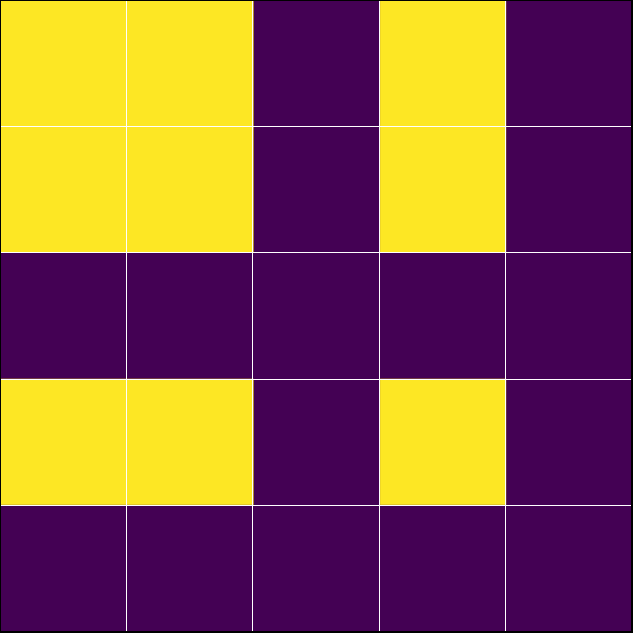}
\end{minipage}%
}\hfil
\subfloat[Cons.\ cluster area\label{fig:csqArea}]{%
\begin{minipage}[t]{.3\textwidth}
  \centering
  \includegraphics[width=.45\textwidth]{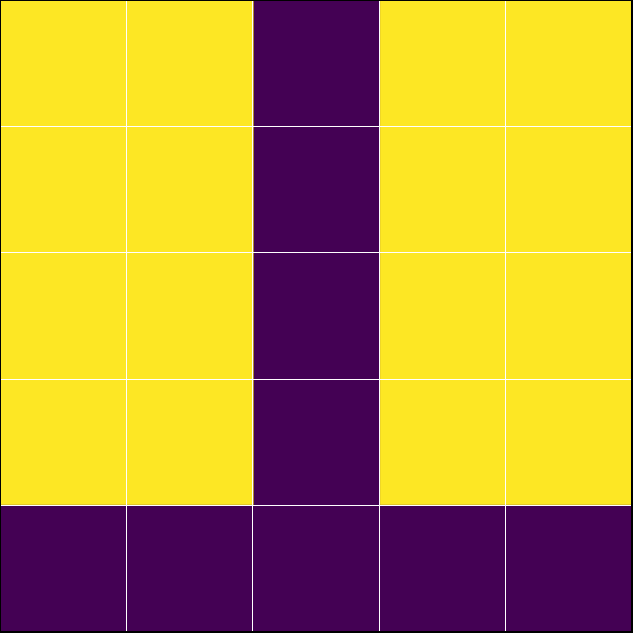}
  \includegraphics[width=.45\textwidth]{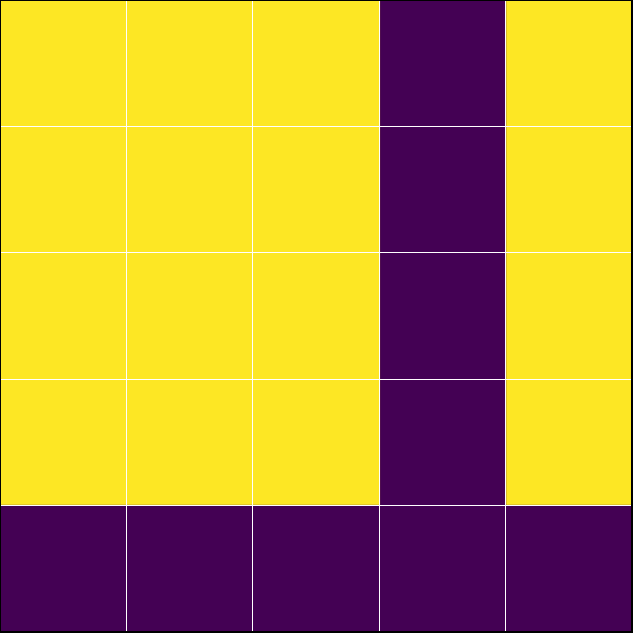}      
\end{minipage}%
}\hfil
\subfloat[Uninterrupted area\label{fig:csqblock}]{%
\begin{minipage}[t]{.3\textwidth}
  \centering
  \includegraphics[width=.45\textwidth]{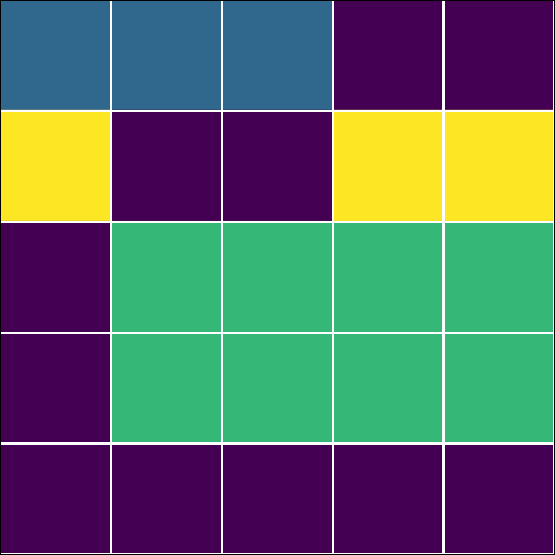}
  \includegraphics[width=.45\textwidth]{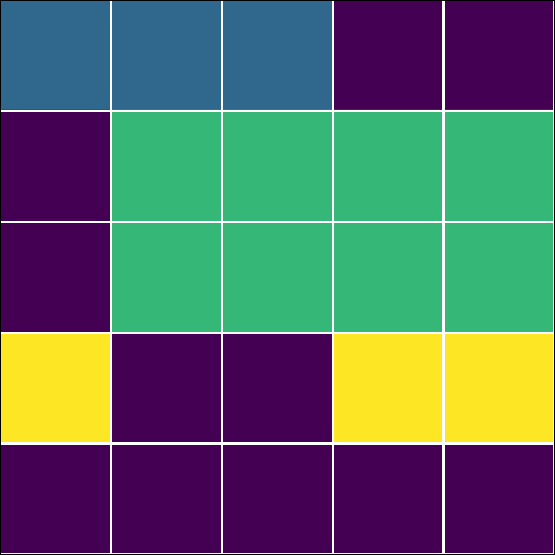}           
\end{minipage}
}%
\caption{Visualizations of biclusters for each of our objective functions. For
	each of them, the right visualization is preferable. Every color represents
	a different bicluster, except for purple which represents 0-elements.
	Observe that in Fig.~\ref{fig:csqblock}, no matter how we arrange the
	columns, one of the three biclusters must always be visualized with
	non-consecutive columns.}
\end{figure}

Recall that biclusters represent pairs of elements that
relate to each other. An ideal depiction of a single
bicluster~$(R_i,C_i)$ consists of a single large consecutive rectangle in
the visualized matrix. More formally, we would like to have that
$\abs{\consR} = 1$ and $\abs{\consC} = 1$.
However, when the row or column clusters of a biclustering overlap, obtaining a
visualization which simultaneously presents all biclusters ideally is not
possible (see, e.g., Fig.~\ref{fig:csqblock}).  Thus, we have to define criteria
that enable us to compare non-ideal depictions of biclusters.

Informally, the three criteria that we study are as follows:
\begin{enumerate}
	\item \textbf{Proximity:} All rows and columns of each bicluster should be
	close to each other, as shown in Fig.~\ref{fig:prox}.
	\item \textbf{Size of the consecutive cluster areas:} The rows and columns of each
	bicluster should form large consecutive areas, as shown in
	Fig.~\ref{fig:csqArea}.
	\item \textbf{Size of uninterrupted areas:}
	Areas that belong to (possibly different) biclusters should form large
	uninterrupted areas, as shown in Fig.~\ref{fig:csqblock}. Unlike the
	previous objectives, this objective is global, i.e., it is not limited to individual biclusters.
\end{enumerate}

\begin{figure}[tbp]
\centering
\subfloat[Convex hull]{%
    \centering
    \includegraphics[height=25mm]{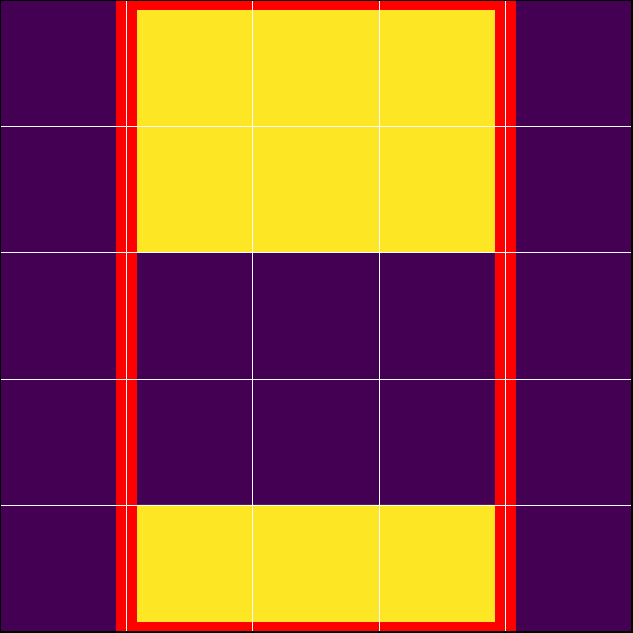}
}\hfil
\subfloat[Example for $\nonzero(\cdot,\cdot)$]{%
\begin{minipage}[t]{.45\textwidth}
         \centering
         \includegraphics[height=30mm]{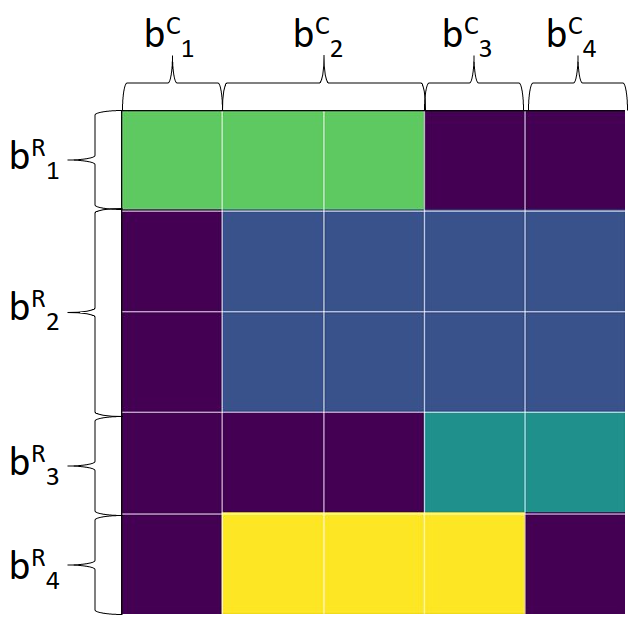}
\end{minipage}
}
\caption{Examples of how the concepts translate to the visualization. We assume $\pi_R= id_R$ and $\pi_C = id_C$. 
Purple colored tiles are 0-elements, and identically colored tiles belong to the same bicluster.
On the left, $\cons(\pi_R(R_1))=\{\{1,2\},\{5\}\}$ and
$\cons(\pi_C(C_1))=\{\{2,3,4\}\}$. The convex hull of the cluster is shown in
red and $S_{\proximity}(\pi_R,\pi_C) =15$. 
On the right, four biclusters are visualized with different color and
 $\nonzero(b^C_2,\pi_R) = \{1,2,3,5\}$ and   $\nonzero(b^C_3,\pi_R) = \{2,3,4,5\}$.
}
\label{fig:definitions}
\end{figure}

The formal definitions follow below. Note that even though the first and second
criteria look similar at first, they are different: even when a bicluster is visualized
with low proximity, it may still consist of several non-consecutive areas.
Furthermore, the third criterion is particularly important when dealing with
non-overlapping biclusterings; it will be useful, for instance, when visualizing
biclusterings that have non-overlapping row clusters but overlapping column
clusters, which is not captured by the previous two definitions.  Previous work
focused on proximity~\cite{jin2008overlapping,colantonio2011visual} and also
implicitly the consecutive area~\cite{colantonio2011visual}.

\label{sec:objectiveFunction}
Next, we formally present three different objective functions, one for each criterion. Having different objective functions, instead of a single combined one, allows a more fine-grained evaluation of the visualizations.

\textbf{Proximity.}
Our first objective function measures proximity. As stated
above, our intuition is that for each bicluster, all of its rows and columns
should be close to each other. To capture this intuition, we want to visualize the biclusters so that the \emph{convex hull} of rows and
columns that belong to the bicluster is small.

Consider permutations $\pi_R$ and $\pi_C$ and a bicluster
$(R_i,C_i)$. The size of the convex hull of 
$(R_i,C_i)$ in the biadjacency matrix $A$ is given by
\begin{equation}
\label{eq:score-proximity}
\begin{split}
&\quad\,S_{\proximity}((R_i, C_i), (\pi_R,\pi_C))\\ 
&= 
    [\max\{\pi_R(R_i)\} - \min\{\pi_R(R_i)\} + 1]
	\cdot [\max\{\pi_C(C_i)\} - \min\{\pi_C(C_i)\} + 1].
\end{split}
 \end{equation}
Observe that for a single bicluster $(R_i,C_i)$ this quantity is minimized when
it is visualized as a single consecutive rectangle, i.e., $\abs{\consR} = 1$ and
$\abs{\consC} = 1$. See also Fig.~\ref{fig:definitions}.

For all $k$~biclusters, our objective function for minimizing the
proximity is
\begin{equation}
\label{eq:proximity}
    f_{\proximity}(\pi_R,\pi_C) = 
	\sum_{i=1}^k
    S_{\proximity}((R_i, C_i), (\pi_R,\pi_C)).
\end{equation}

\textbf{Size of the consecutive cluster areas.}
Next, we consider the aspect that the rows and columns of the same bicluster
should form large consecutive areas.

First observe that when we visualize
a bicluster $(R_i,C_i)$ under permutations $\pi_R$ and $\pi_C$, then the
consecutive areas are given by $\consR \times \consC$. For instance, if
$\consR = \{ \{1,2\}, \{5\} \}$ and $\consC = \{ \{3\}, \{7\} \}$, then the
consecutive areas are  $\{\{(1,3),(2,3)\}, \{(5,3)\}, \{(1,7),(2,7)\},
\{(5,7)\} \}$.

Given this observation, we define the score $S_{\clusterArea}((R_i,C_i),(\pi_R,\pi_C))$ for a
single bicluster $(R_i,C_i)$ under the permutations $\pi_R$ and $\pi_C$ as
follows:
\begin{equation}
\label{eq:score-area}
	S_{\clusterArea}((R_i,C_i), (\pi_R, \pi_C)) = \sum_{(X,Y) \in \consR\times\consC} \abs{X \times Y}^2.
\end{equation}

In this score, we sum over the \emph{squared} areas of the induced submatrices.
Note that maximizing this score incentivizes layouts with larger consecutive
areas.  In particular, $S_{\clusterArea}((R_i,C_i), (\pi_R,\pi_C))$ is maximized iff the
bicluster $(R_i,C_i)$ is visualized as a single connected component, i.e., when
the rows and columns in $\Rreordered$ and $\Creordered$ are consecutive.
Also observe that if in the score we summed over $\abs{X \times Y}$ instead
of $\abs{X \times Y}^2$,  the sum would be independent of the permutations
and always equal to $\abs{R_i \times C_i}$; this is why we sum over $\abs{X \times Y}^2$.

The corresponding global objective function is:
\begin{align}
\label{eq:cluster-area}
    f_{\clusterArea}(\pi_R,\pi_C) =
	\sum_{i=1}^{k} S_{\clusterArea}((R_i,C_i),(\pi_R,\pi_C)).
\end{align}

\textbf{Size of uninterrupted areas.}
Lastly, we introduce an objective function which incentivizes that
areas that belong to (possibly different) biclusters should form large
uninterrupted areas.
This is for useful for visualizing biclusters that are similar but
non-overlapping, e.g., because they have disjoint row clusters but highly
similar column clusters. %

Recall that $\mathcal{B}^R=\{b_1^R,\dots,b_s^R\}$ and
$\mathcal{B}^C=\{b_1^C,\dots,b_t^C\}$ are the row and column blocks,
respectively.  Since splitting up elements of blocks would only be detrimental
to our visualizations, we henceforth assume that the elements from all row and
column blocks are consecutive in our permutations, i.e.,
$\abs{\cons(\pi_R(b_i^R))} = 1$ and $\abs{\cons(\pi_C(b_j^C))} = 1$ for all $i$
and $j$.

Now consider a row block $b_i^R$ and the submatrix $A[\pi_R(b_i^R),:]$ which it
induces. Observe that in this submatrix,
column~$\pi_C(c)$ is contained in a bicluster if $c\in b_j^C$ and
$\clusters_R(b_i^R)\cap\clusters_C(b_j^C) \neq \emptyset$, i.e., if $c$ is from
a column block~$b_j^C$ which co-occurs in a bicluster together with a row block
$b_i^R$. Similarly, if $c\in b_j^R$ for $j$ with
$\clusters_R(b_i^R)\cap\clusters_C(b_j^C)=\emptyset$ then column
$\pi_C(c)$ is not contained in a bicluster. Thus,  the set of
all columns in $A[\pi_R(b_i^R),:]$ which are contained a bicluster after
applying the permutation $\pi_C$ is
$\nonzero(b_i^R,\pi_C) := \bigcup_{j \colon \clusters_R(b_i^R) \cap \clusters_C(b_j^C) \neq \emptyset} \pi_C(b_j^C)$. See Fig.~\ref{fig:definitions} for an example.
Thus, the size of the uninterrupted area of columns in biclusters in
$A[\pi_R(b_i^R),:]$ is given by
\begin{equation}
\label{eq:block-slice-row}
	S_{\uninterrupted}^R(b_i^R, \pi_C) = 
	\sum_{Y\in\cons(\nonzero(b_i^R,\pi_C))}
	\abs{b_i^R \times Y}^2.
\end{equation}

Notice the similarity of~\eqref{eq:block-slice-row} and \eqref{eq:score-area}
above. The main difference is that in \eqref{eq:score-area} we sum over
the areas induced by the biclusters, whereas here we sum over the area induced by columns inside biclusters, regardless of bicluster membership. This is beneficial since when
row block $b_i^R$ co-occurs with column block $b_{j_1}^C$ and with column block
$b_{j_2}^C$, then this definition incentivizes to place the column blocks
$b_{j_1}^C$ and $b_{j_2}^C$ next to each other, even though they might not share
a bicluster (i.e., even when $\clusters_C(b_{j_1}^C) \cap \clusters_C(b_{j_2}^C)
= \emptyset$), which is not covered by \eqref{eq:score-area}.

Similar to above, we also want to measure the area of consecutive rows that appear in a bicluster in
a submatrix $A[:,\pi_C(b_j^C)]$ that is induced by a fixed column cluster
$b_j^C$. We thus define
$\nonzero(b_j^C,\pi_R) = \bigcup_{i \colon \clusters_R(b_i^R) \cap \clusters_C(b_j^C) \neq \emptyset}
	\pi_R(b_i^R)$ and set
	$S_{\uninterrupted}^C(b_j^C, \pi_R) = 
	\sum_{X\in\cons(\nonzero(b_j^R,\pi_R))}
	\abs{X \times b_j^C}^2$.

Now our overall objective function becomes:
\begin{equation}
\label{eq:blockSliceScore}
    f_{\uninterrupted}(\pi_R,\pi_C) = 
	\sum_{i=1}^s S_{\uninterrupted}^R(b_i^R, \pi_C)
	+ \sum_{j=1}^t S_{\uninterrupted}^C(b_j^C, \pi_R).
\end{equation}

\section{Algorithms}

In this section, we describe our algorithms to obtain the
permutations $\pi_R \colon [m] \to [m]$ and $\pi_C \colon [n] \to [n]$
that optimize our objective functions.

For better efficiency, we focus on finding permutations in
which the rows of row blocks are always consecutive (and the same holds for the
columns of column blocks).  Observe that this assumption is without loss of
generality, i.e., splitting the elements of a row or column block into multiple
consecutive parts will never improve the objective functions we study.

Thus, suppose that we have row blocks $b^R_1,\dots,b^R_s\subseteq[m]$. Then our
new goal is to find a row block permutation $\sigma\colon[s]\to[s]$ that
optimizes our objective functions.\footnote{
Note that we can turn the row block permutation $\sigma$ into a row permutation
$\pi_R\colon[m]\to[m]$ as follows: For each $i\in[s]$, we fix an
arbitrary order of the elements in $b^R_i$. Now we create a list~$L$ by
iterating over $i\in[s]$ and adding the elements in $b^R_{\sigma(i)}$ one
after another to~$L$.  If element $r$ is at the $p$'th position in $L$, then we set
$\pi_R(r)=p$.
} This will be more efficient since in practice
$s\ll m$. The same can be done for finding a column block permutation.

We present the pseudocode of our algorithms in Appendix~\ref{sec:pseudocode}.

\subsection{Greedy Algorithms}
\label{sec:greedy}

We start by considering a simple greedy algorithm for optimizing the three
objective functions from Sect.~\ref{sec:objectiveFunction}.
Our algorithm starts by sorting the row and column blocks based on their
importance. Here, the \emph{importance score} of a block~$b$ is the sum of the
area of the biclusters $b$ belongs to, i.e., 
$\sum_{i\in\clusters(b)} \abs{R_i \times C_i}$.  The idea is that blocks which
are involved in large clusters are treated first and thus have priority when
picking their position.

The greedy algorithm computes the row and column block permutations $\sigma_R$
and $\sigma_C$ simultaneously. Initially, they are set to the empty permutations
$\sigma_R\gets \emptyset$ and $\sigma_C\gets \emptyset$ without any elements.
Now the greedy algorithm proceeds in iterations until all row and column blocks
have been assigned to the permutations. In iteration~$j$, we add the row
(column) block~$b$ with $j$'th highest importance to
$\sigma_R$ ($\sigma_C$). To pick the position of~$b$, we iterate over 
$i=1,\dots,j$ and consider the permutation $\sigma_R$
with $b$ added in the $i$'th position. Then we insert $b$ in the position~$i^*$ that
achieved the best objective function value.

We note that while building the permutations above, they only map to the subset
of the rows and columns that are contained in the row and column blocks that
were assigned to the permutations. Therefore, to compute the objective function
values, we only consider rows and columns that are contained in blocks that were
already added to the permutations. Details and pseudocode are available in Appendix~\ref{sec:pseudocode-greedy}.

\subsection{Demerit-Based Algorithms}
\label{sec:demerit}

In practice, the greedy
algorithm can be inefficient as it recomputes the objective functions several times
during each iteration and each such recomputation requires a global pass over all
clusters and blocks.

To remedy this problem, next we introduce the notion of \textit{demerit},
which can be optimized locally and which acts as a penalty function for placing
dissimilar blocks next to each other.
Formally, the \emph{demerit for row block~$b^R$ and column blocks $b_i^C$ and $b_j^C$} is
given by:
\begin{equation*}
	\demerit(b^R; b_i^C, b_j^C)
	= \begin{cases} 
      \abs{b^R}\cdot(\abs{c_1 \cup c_2} +1) & \text{if } c_1 = \emptyset \text{ or } c_2 = \emptyset,  \\
      \abs{b^R}\cdot(\abs{c_1 \cup c_2} - \abs{c_1 \cap c_2})  & \text{otherwise},
   \end{cases}
\end{equation*}
where $c_1 = \clusters_R(b^R) \cap \clusters_C(b_i^C)$
and $c_2 = \clusters_R(b^R) \cap \clusters_C(b^C_j)$.
Observe that the demerit is a penalty term that measures the size of the row block $b^R$ and how dissimilar the
blocks $b_i^C$ and $b_j^C$ are in terms of their cluster membership, i.e., it
counts the number of clusters which contain row block~$b^R$ but only exactly one
of $b_i^C$ and $b_j^C$.

To measure the \emph{demerit of the column block permutation~$\sigma_C$}, we
set
\begin{align*}
	\demerit(\sigma_C)
	= \sum_{b^R \in \mathcal{B}^R} \sum_{i=1}^{t-1}
		\demerit(b^R; b_{\sigma_C(i)}^C, b_{\sigma_C(i+1)}^C),
\end{align*}
where $t$ is the number of column blocks. This is the overall penalty
incurred across all row blocks for column blocks that are placed next to each
other. Optimizing this objective function should be somewhat
simpler than the previous ones, because we only have to consider consecutive
pairs of column blocks $b_{\sigma_C(i)}^C$ and $b_{\sigma_C(i+1)}^C$, which can
be checked locally. This is in contrast to our previous objective functions,
which have to globally take into account all blocks that belong to a single
bicluster (proximity and consecutive cluster area) or all blocks that appear
consecutively (uninterrupted area).

Next, we introduce algorithms for minimizing the demerit, where we assume that we have a fixed
row block permutation $\sigma_R$ and we wish to compute an improved ordering
of the column blocks $\sigma_C$.
The same procedure can be used for fixed~$\sigma_C$ and for finding~$\sigma_R$ with small demerit. Details of both algorithms are available in Appendix~\ref{sec:greedy-demerit}.

\textbf{TSP heuristic.}
We first consider a TSP (traveling salesperson) heuristic to find a permutation~$\sigma_C$ that
minimizes the demerit. First, we construct a complete graph 
containing all column blocks $b^C\in\mathcal{B}^C$ as nodes. For two column
blocks~$b_i^C$ and~$b_j^C$, we set the weight of the corresponding edge to
	$w_{i,j} = 
	\sum_{b^R \in \mathcal{B}^R} 
	\demerit(b^R; b_i^C, b_j^C)$
which corresponds to the demerit of placing $b_i^C$ and $b_j^C$ next to each
other. This is a complete graph, i.e., there are edges for
all pairs of column blocks. Then we use a TSP solver to find a TSP tour in the corresponding graph, which is given by a
cycle~$(b_{i_1}^C,\dots,b_{i_t}^C)$ that visits every vertex exactly once. This
corresponds to a column block permutation~$\sigma_C$. Since the objective of TSP
is to minimize the cost of the cycle, this corresponds to minimizing the
demerit. Note that for defining $\sigma_C$, we can start with any of the blocks
from the cycle, i.e., we can set $\sigma_C(1) = b_{i_j}^C$ for any $j$ and then
proceed in the order of the cycle. To obtain the best results in practice, we
pick the value~$j$ which maximizes the cluster area \eqref{eq:cluster-area}.

\textbf{Greedy demerit algorithm.} We also consider a greedy algorithm which orders the
blocks by their importance score and inserts them one by one.
When inserting a block, it tries out all possible positions and picks the one
which minimizes the total demerit.

\subsection{Post-Processing: Suggesting Unclustered Rows and Columns}
\label{sec:post-processing}

Finally, we present a post-processing scheme that finds unclustered rows
and columns that have high similarity with existing biclusters. This will enable
domain experts to easily identify structures which might have been missed by the
biclustering algorithm. We describe our post-processing scheme for finding
unclustered rows whose 1-entries have high similarity to existing column
clusters; it can also be used for finding columns that are similar to
existing row clusters.

We say that a row~$r$ is \emph{unclustered} if it is not contained in any
row cluster, i.e., if $r\not\in \bigcup_i R_i$, and we write $\bar{R}$ to denote the
set of unclustered rows.  For $r\in \bar{R}$, we write $\nonzero(r)=\{c \in C :
A_{rc} = 1\}$ to denote the columns of all 1-entries in~$r$. Now the
\emph{similarity} of $r$ and a column cluster~$C_i$ is given by
$\similarity(r,C_i) = \abs{C_i}^{-1} \abs{\nonzero(r) \cap C_i}$, i.e., it measures the fraction of elements from $C_i$ that also appear in $\nonzero(r)$.
Furthermore, the \emph{density of a bicluster $(R_i,C_i)$} is 
$\density(R_i,C_i)=(\abs{R_i}\cdot\abs{C_i})^{-1}\sum_{r\in R_i, c\in C_i}
A_{r,c}$, i.e., it is the average number of non-zero entries in the submatrix
induced by $R_i \times C_i$.

Now our idea is to create biclusters $(\bar{R}_i,C_i)$, which consist of unclustered
rows~$\bar{R}_i$ and ``original'' column clusters~$C_i$. Here, we assign a row
$r\in \bar{R}$ to $\bar{R}_i$ if $\similarity(r,C_i) \geq \density(R_i,C_i)/2$.
This encodes the intuition that the rows in $\bar{R}_i$ are allowed to be slightly sparser
than those in the original bicluster $(R_i,C_i)$; for a domain expert it
might be interesting to inspect them because the original biclustering algorithm
might have ``missed'' them.

In the visualization, these new biclusters have a special place. The original biclusters~$(R_i,C_i)$ are situated in the middle of the figure. Then, adjacent to that central part, the new biclusters~$(\bar{R}_i,C_i)$ and $(R_i,\bar{C}_i)$ are added, and then the remaining unclustered rows and columns follow.

\section{Experiments}

We implemented our algorithms in Python and we practically evaluate them on
real-world datasets. The source code\footref{fn:code} and the
plots\footref{fn:plots} of all biclusterings are
available on GitHub. The
experiments were performed on a 40-core Intel(R) Xeon(R) CPU E5-2630 v4
@ 2.20GHz.

The datasets we used are listed in Table~\ref{table:experiments}, where we
focussed on small- to medium-sized datasets since visualizing very large
datasets requires other techniques~\cite{pezzotti18multiscale}.  We note that
for 20news and movieLens we only considered the top-$500$ densest rows and
columns to reduce the size of the datasets.

\begin{wraptable}[16]{R}{0.55\textwidth}
\centering
\begin{threeparttable}
 \caption{Datasets used in the experiments}
  \label{table:experiments}
  \small
\begin{tabular}{
@{}
l
S[table-format=5.0,group-minimum-digits=3]
S[table-format=5.0,group-minimum-digits=3]
S[table-format=1.3,round-mode=places,round-precision=3]
l
@{}
} 
 \toprule
 dataset & {rows} & {columns} & {density} & {ref.} \\ 
 \midrule
 20news & 500  & 500& \fpeval{55367/(500*500)} & \cite{TwentyNews} \\ 
 americas\_large &  3485 &10127 &\fpeval{185294/(3485*10127)} & \cite{molloy2009evaluating}\\ 
 americas\_small &3477 &1687 &\fpeval{105205/(3477*1687)} & \cite{molloy2009evaluating}\\
 apj & 2044 &1164 & \fpeval{6841/(2044*1164)}& \cite{molloy2009evaluating}\\
dialect & 1334 & 506 &\fpeval{108932/(1334*506)} & \cite{embleton97finnish,embleton00computerized}\\
domino & 79 &231  &\fpeval{730/(79*231)}& \cite{molloy2009evaluating}\\
fire1 &  365 & 709 & \fpeval{31951/(365*709)} & \cite{molloy2009evaluating}\\
fire2 &  325 & 709  & \fpeval{36428/(325*709)}& \cite{molloy2009evaluating}\\
healthcare   & 46& 46& \fpeval{1486/(46*46)}& \cite{molloy2009evaluating} \\
movieLens  &500&500 & \fpeval{137569/(500*500)} & \cite{scikit-learn} \\
Mushroom (sample) & 250& 117 &\fpeval{10750/(250*117)} & \cite{jin2008overlapping} \\
paleo & 124&139 & \fpeval{1978/(124*139)} & \cite{NOWData}  \\
\bottomrule
\end{tabular}
\end{threeparttable}
\end{wraptable}

To obtain our biclusterings, we used the PCV algorithm
\cite{neumann18bipartite}, which returns non-overlapping row clusters but
overlapping column clusters, and the basso algorithm
\cite{miettinen08discrete}, which returns overlapping row and column clusters.
Both algorithms have a parameter $k$ that determines the number of clusters and
we report the choice of $k$ for each experiment.

\begin{figure}[t]
\centering
\centering
\subfloat[ADVISER\label{20newsk10ADVISER}]{%
\begin{minipage}{.3\textwidth}
\centering
\includegraphics[height=50mm]{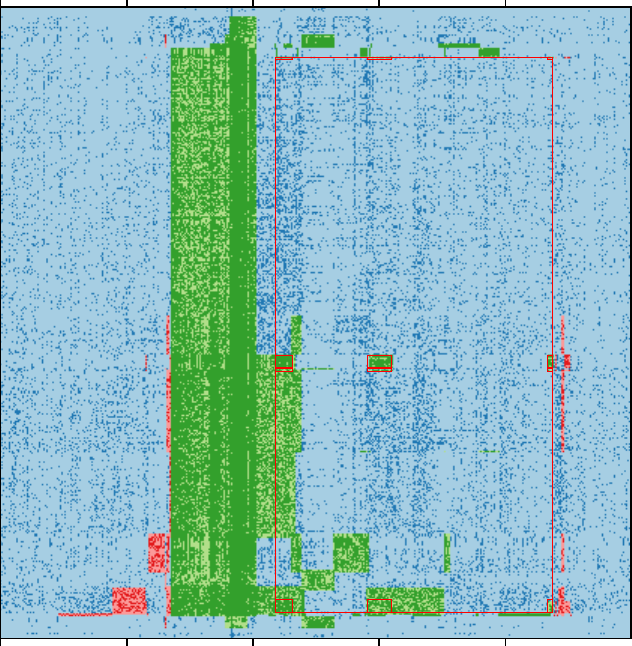}
\end{minipage}%
}\hfil
\subfloat[greedyProximity\label{fire1basso5Proximity}]{%
\begin{minipage}{.6\textwidth}
\centering
\includegraphics[height=50mm]{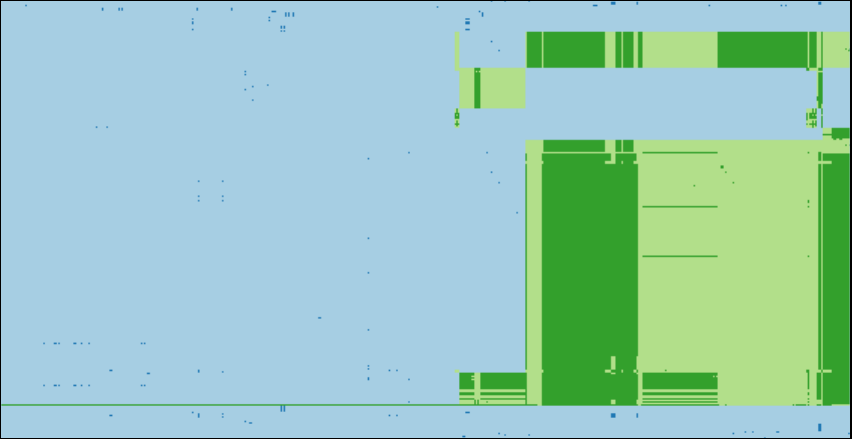}
\end{minipage}%
}\\
\subfloat[TSPheuristic\label{20newsk10TSP}]{%
\begin{minipage}{.3\textwidth}
\centering
\includegraphics[height=50mm]{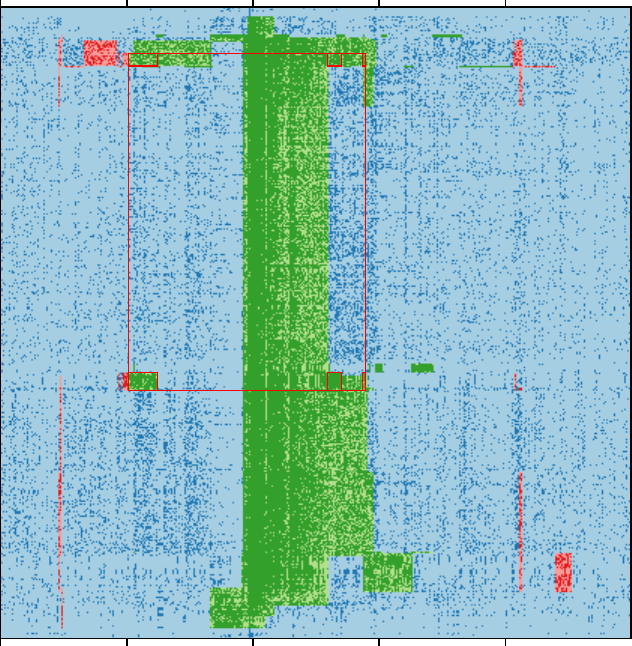}
\end{minipage}%
}\hfil
\subfloat[greedyDemerit\label{fire1basso5Demerit}]{%
\begin{minipage}{.6\textwidth}
\centering
\includegraphics[height=50mm]{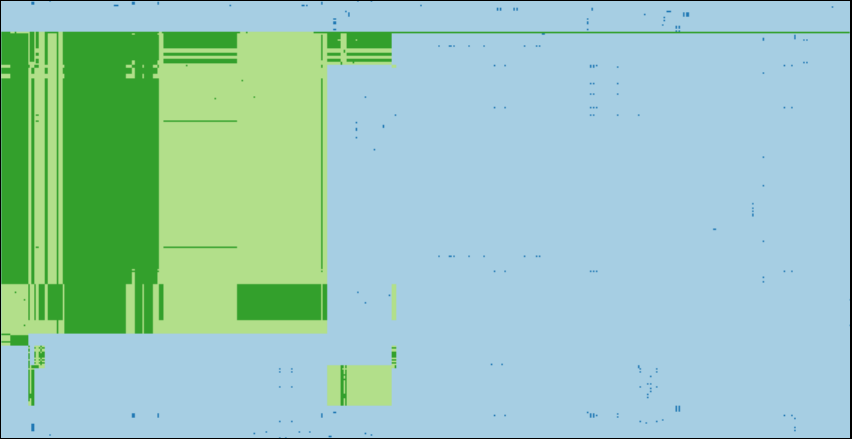}
\end{minipage}%
}
\caption{Visualizations (a) and (c) represent 20news. The red boxes denote
		the convex hull of the same cluster in both visualizations. The
		biclustering was obtained using basso with $k=11$. Visualizations (b) and (d) represent Fire1.
	The biclustering was obtained using basso with $k=10$.}
\end{figure}

In some of our visualizations, we use a 6-color system to convey more
information (e.g., Fig.~\ref{fig:beat-adviser},~\ref{20newsk10ADVISER} and~\ref{20newsk10TSP}).  Each of
the colors is associated with a distinct category of data in the visualization:
clustered elements appear in green, unclustered elements that were picked in our
post-processing step (Sect.~\ref{sec:post-processing}) are red, and all
remaining unclustered elements are blue.  The dark tones of each color
correspond to $1$-entries in the original matrix.\footnote{We picked the colors using color
	brewer~\cite{brewer2009colorbrewer}, so that the core set of colors (excluding
	the post-processing step) is colorblind safe and print friendly.
	As there is no 6-colors set that is colorblind safe, the
	final set of colors only retains the print friendly property.}
This allows us to assert whether or not an element belongs to the biclustering
and/or to the original data, and it further allows to assess the density of the
clustered (and non-clustered) areas.

In our experiments, we consider four greedy algorithms for optimizing the
objective functions, denoting them greedyProximity,
greedyConsecutiveClustersArea, greedyUninterruptedArea, greedyDemerit. Our
TSP-based algorithm is denoted TSPheuristic and
to solve TSP we use a solver from Google OR tools~\cite{ortools}.  We compare them against
the state-of-the-art method ADVISER~\cite{colantonio2011visual}. Since there was no code available for ADVISER, we implemented our own version of it, available with our software.

\textbf{Qualitative Evaluation.}

\begin{figure}[t]
\centering
\subfloat[Effect of reordering\label{fire1basso5}]{%
\begin{minipage}{.3\textwidth}
\centering
\includegraphics[angle=90, height=.9\textwidth]{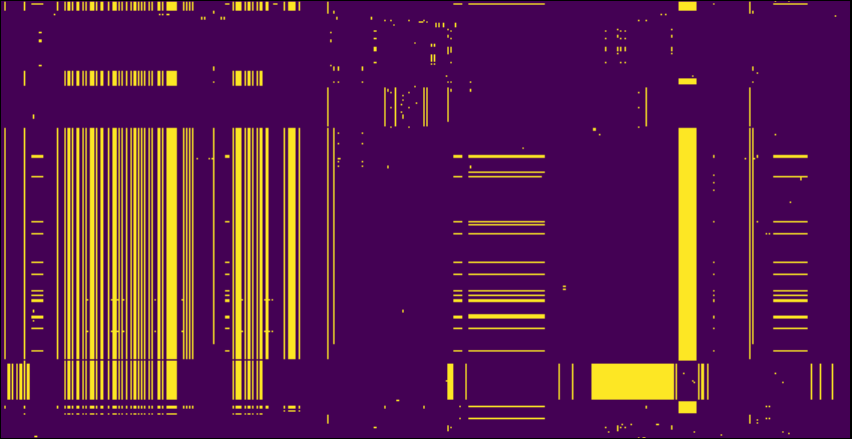}
\includegraphics[angle=90, height=.9\textwidth]{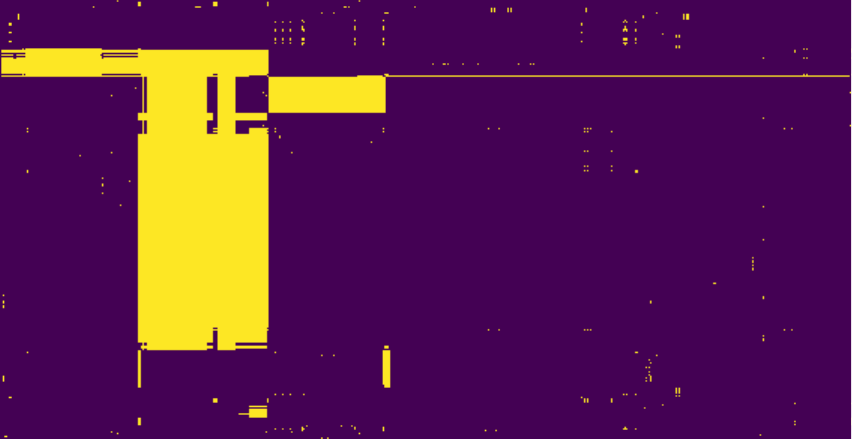}  
\end{minipage}%
}%
\hfil
\subfloat[Effect of color scheme\label{fig:orphans}]{%
\begin{minipage}{.68\textwidth}
\centering
\includegraphics[trim={3.6cm 1.2cm 2.8cm 1.3cm}, clip, width=.45\textwidth]{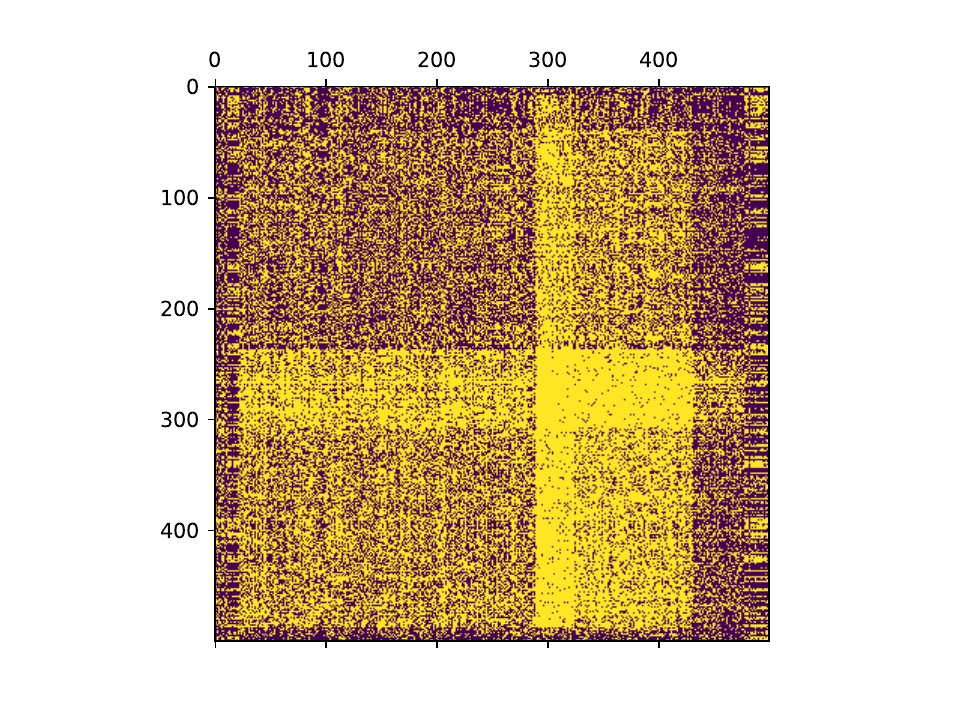}
\includegraphics[trim={3.6cm 1.2cm 2.8cm 1.3cm}, clip, width=.45\textwidth]{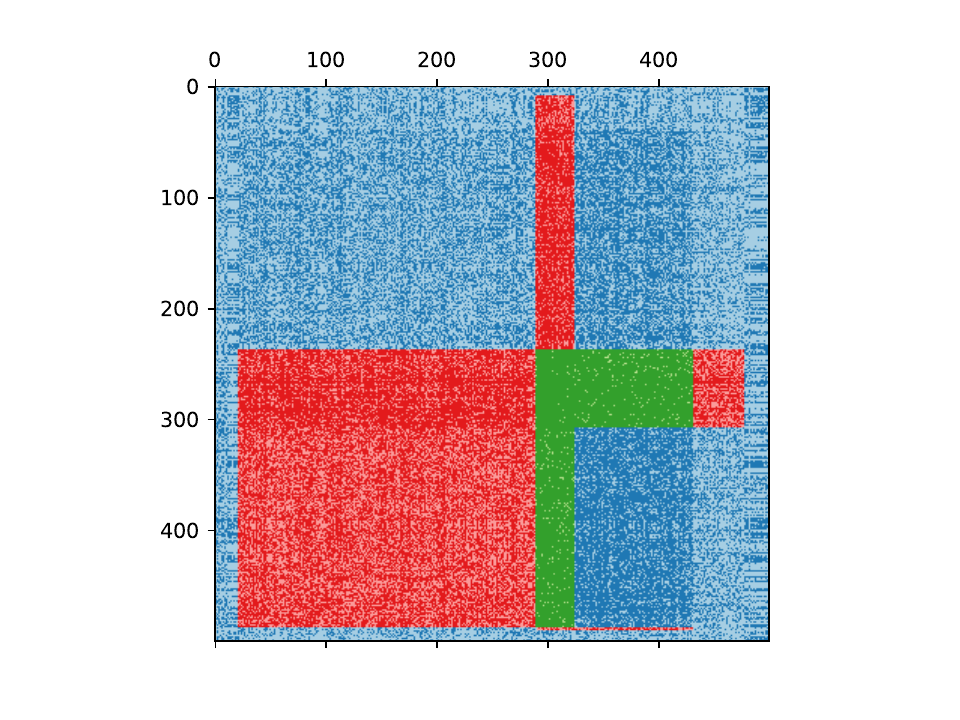}
\end{minipage}%
}
\caption{(a): Visualization of Fire1 without any reordering (left) and after reording using TSPheuristic (right).
 (b): Visualization of movieLens using ADVISER. The result is colored using plain 0/1 entries (left) and our color scheme (right).
}
\end{figure}
We start with the qualitative evaluation of the algorithms.  Our findings in this
section are twofold: our TSPheuristic provides better visualizations than
ADVISER~\cite{colantonio2011visual} and our objective functions indeed measure
the aspects of the visualizations which they are supposed to measure.

First, let us briefly argue about the merit of visualizing biclusterings. In
Fig.~\ref{fire1basso5}, we present visualizations of Fire1 without
any ordering and the visualization created using TSPheuristic. The unordered dataset hints that some
rows and columns seem related. After using basso for biclustering with $k=5$ and reordering
the data with TSPheuristic, we can easily see the relation between rows and
columns, as well as notice sparser areas inside the biclusters.

Next, we consider biclusterings obtained from PCV, which returns
non-overlapping row clusters but overlapping column clusters.
Fig.s~\ref{dialectPCV48} and~\ref{paleo50} depict visualizations of
dialect and paleo using ADVISER and TSPheuristic.  Since ADVISER's
objective function does not take into account uninterrupted areas, its
visualization is much less coherent than the one by TSPheuristic.
The uninterrupted areas objective function captures this
aspect well, where TSPheuristic obtains an \SI{18.9}{\percent} higher score on dialect
and a \SI{7}{\percent} higher score on paleo.

Now we consider basso's more complex biclusterings for 20news with $k=11$, which contains overlapping row and
column clusters. In this case, TSPheuristic is more
resilient than ADVISER w.r.t.\ the proximity of the bicluster elements.
In Fig.s~\ref{20newsk10ADVISER} and~\ref{20newsk10TSP}, we show the convex hulls of the same bicluster in
red. One can see that the representation of the cluster is more compact in
the visualization generated from TSPheuristic, compared to ADVISER. This also
translates to the proximity objective function, where TSPheuristic achieves a
proximity score that is \SI{30.8}{\percent} lower than that of ADVISER (note that 
optimizing the proximity is a minimization problem).

Next, let us consider the uninterrupted areas that are generated by the
algorithms. In Fig.s~\ref{fire1basso5Proximity} and~\ref{fire1basso5Demerit}, the respective biclusters have a
similar proximity score (\num{269125} and \num{253400}), but the visualization proposed by
greedyDemerit is nicer, as all the clusters are drawn as one consecutive block.
This is also highlighted by the objective function value for uninterrupted
bicluster area, which is \SI{26.4}{\percent} higher for greedyDemerit.

Finally, we highlight the usefulness of our coloring scheme and our
post-processing step from Sect.~\ref{sec:post-processing} in
Fig.~\ref{fig:orphans}. The coloring scheme highlights the very dense areas
that basso selected as biclusters in green. Then our post-processing scheme
clearly indicates that the remaining unclustered rows and columns contain areas
similar to the original clusters, but of slightly lower density, which could be
worth considering when manually inspecting the clusters. We note that the dense
area in the bottom right of the plot is not marked in red, since the
corresponding submatrix was not considered as part of the bicluster by basso;
we decided not to consider such areas in our post-processing step.

\textbf{Quantitative Evaluation.}

\begin{figure}[t]
    \centering
    \subfloat[Results on biclusterings computed by the PCV algorithm\label{fig:objectivePCV}]{%
    \begin{minipage}{.7\textwidth}
    \includegraphics[width=\textwidth]{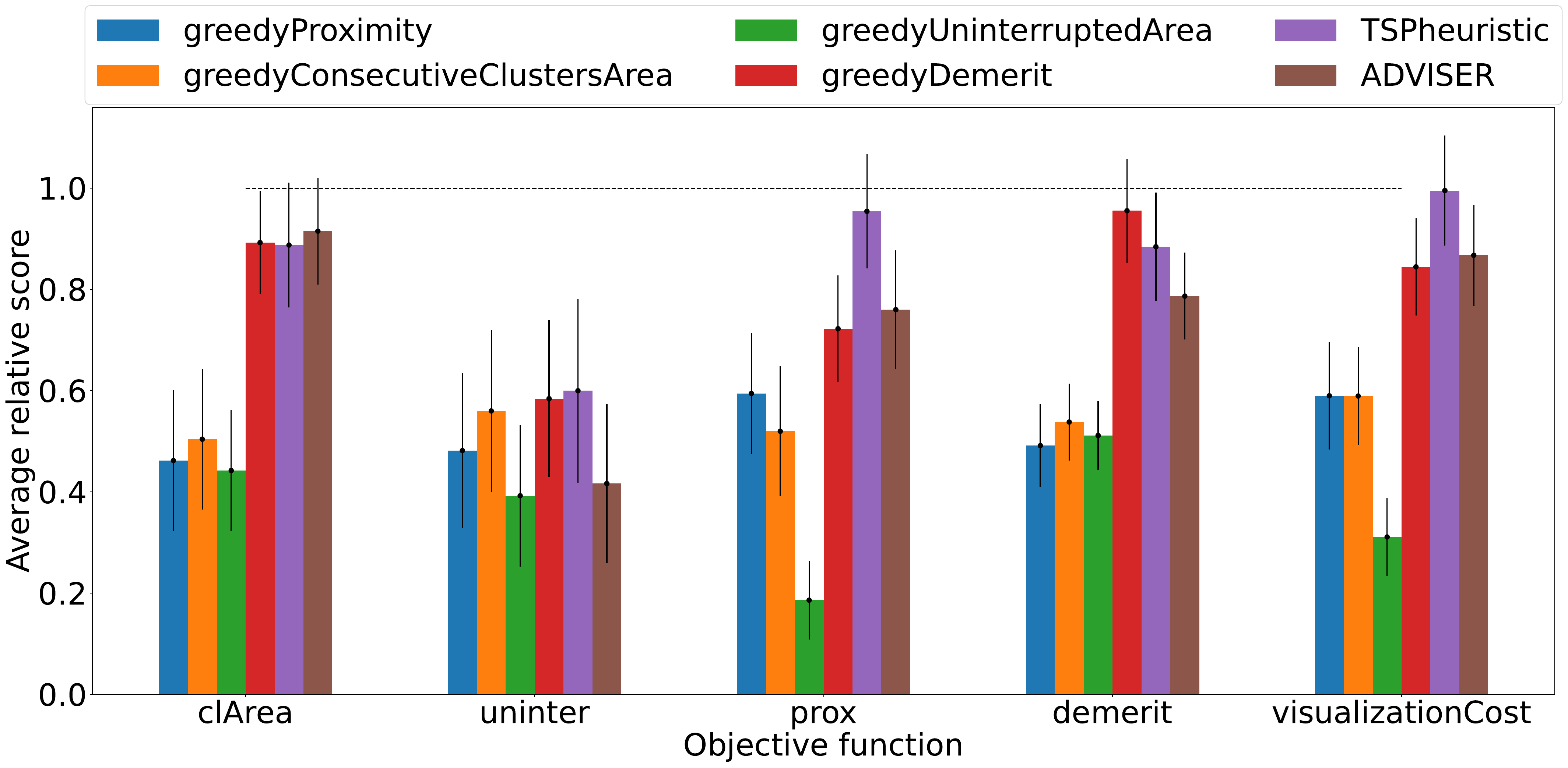}%
    \end{minipage}%
    }\hfil
    \subfloat[Results on biclusterings computed by the basso algorithm\label{fig:objectivebasso}]{%
    \begin{minipage}{.7\textwidth}
    \includegraphics[width=\textwidth]{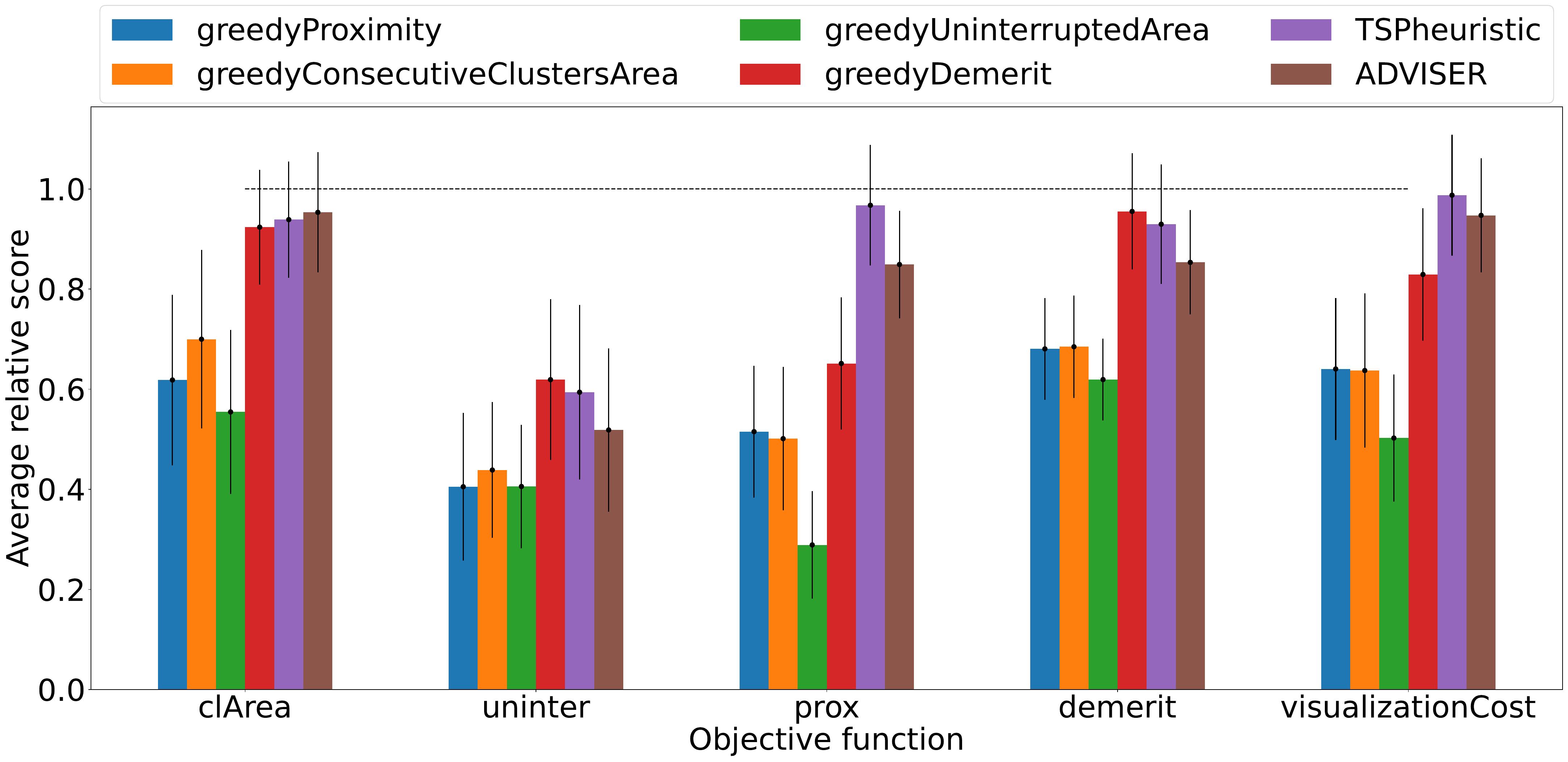}
    \end{minipage}%
    }
    \caption{Aggregated ratio values, grouped by the clustering algorithm used. The reported numbers are averages over all datasets and all $k$, error bars are the variances of the ratio values.}
    \label{fig:objectiveValues}
\end{figure}
For the quantitative validation, we run basso and PCV on all datasets from
Table~\ref{table:experiments} with $k=6,10,14,\dots,54$ to obtain biclusterings.
We run all visualization algorithms for each of these
biclusterings and compute our objective function values from
Sect.s~\ref{sec:objectiveFunction} and~\ref{sec:demerit}. We also compute the
objective function \emph{visualisationCost} from
ADVISER~\cite{colantonio2011visual}. All plots are available
online.\footref{fn:plots}

To obtain comparability across different datasets and different biclusterings,
we use normalization: For each dataset and a fixed biclustering, we
report the ratio 
$r_{A}^{f}=\frac{f(A)-\averageRandomScore}{\max_{A'\in \mathcal{A}} (f(A'))-\averageRandomScore}$,
where $A$ is the visualization algorithm we consider, $f(A)$ is the objective
function value obtained by~$A$ and $\mathcal{A}$ is the set of all
visualization algorithms.
We subtract the $\averageRandomScore$ which denotes the average objective
function of five random permutations; this is motivated by the fact that even
the worst possible visualization will achieve non-negligible scores in our
objective functions since typically their values are lower bounded by the
squares of the block sizes.
Note that if $r_{A}^{f} = 1$, $A$ achieved the best objective function value
among all algorithms we compare.

We report our experimental results in Fig.~\ref{fig:objectiveValues}, where Fig.~\ref{fig:objectivePCV}
presents the results on biclusterings that were generated by PCV and
Fig.~\ref{fig:objectivebasso} presents the results on biclusterings that were generated by
basso. We observe that TSPheuristic performs well across all objective
functions, even though on the consecutiveClusterArea it is slightly outperformed
by ADVISER. Notably, TSPheuristic performs significantly better for the
uninterrupted area score compared to ADVISER, especially for the biclusterings
that were computed by PCV; this corroborates our findings from the qualitative evaluation. We note that among the greedy algorithms, greedyDemerit is the best, which further underscores that using demerit to guide visualizations is a
good idea.  Furthermore, on both sets of experiments, TSPheuristic outperforms
ADVISER on the \emph{visualisationCost} objective function which is being optimized by
ADVISER. We conclude that TSPheuristic provides the best tradeoff across the different
datasets and objective functions.

\textbf{Running Time Analysis.}

\begin{figure}[t]
    \centering
    \includegraphics[width=.7\linewidth]{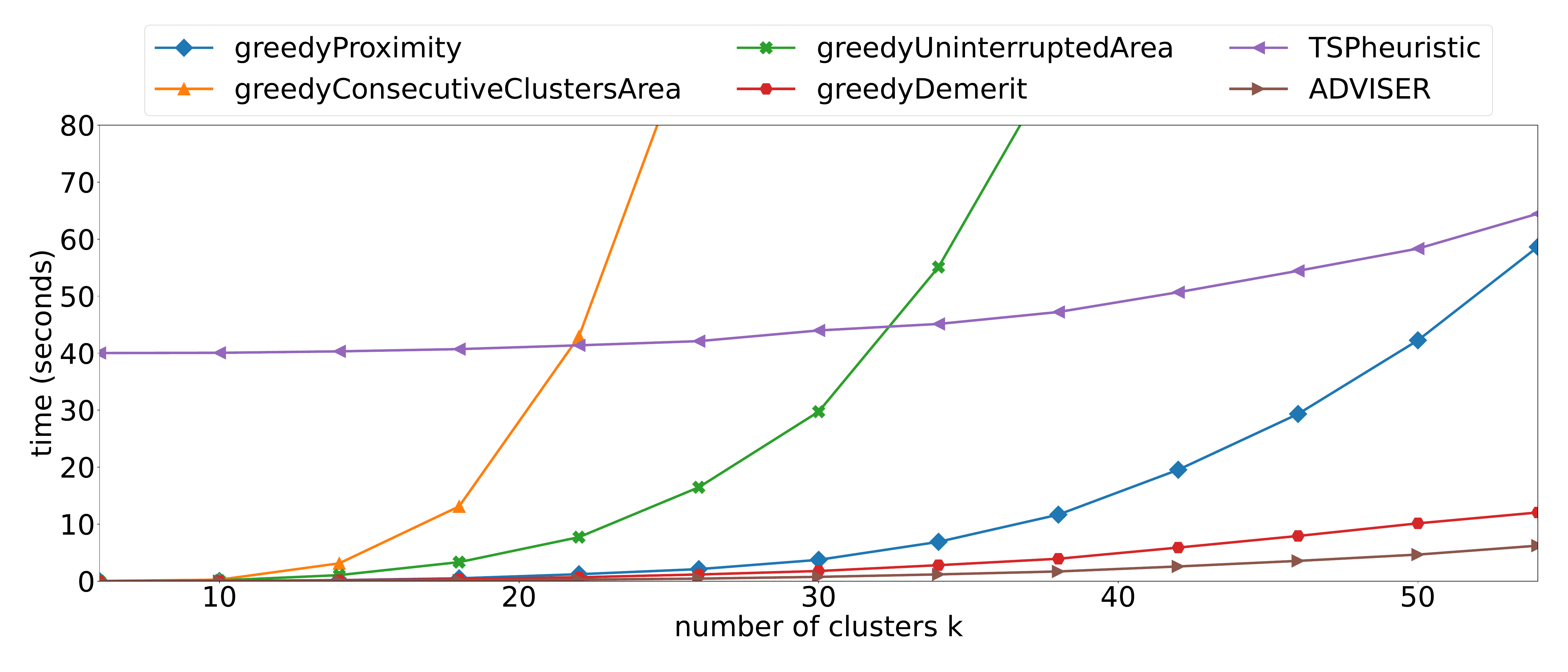}
    \caption{Scalability on americas\_large for varying numbers of clusters~$k$. %
    }
    \label{fig:scalability}
\end{figure}
Finally, we compare the running times of the algorithm on our largest dataset
americas\_large and report the running times in Fig.~\ref{fig:scalability}.
We observe that ADVISER is the fastest method overall and the greedy algorithms
which optimize our objective functions from Sec.~\ref{sec:objectiveFunction}
are slow since computing the objective functions is rather slow. For
TSPheuristic, we see an offset of 40~seconds which is the time we spend on
running the TSP solver; overall, it scales better than our greedy baselines 
and typically finishes in at most 1~minute.

\section{Conclusion}
We studied the visualization of overlapping biclusterings and identified three different aspects that good visualizations should satisfy: 
proximity of cluster elements, large consecutive areas consisting of cluster elements, and large uninterrupted areas of clusters.
We provided objective functions that capture these goals and showed experimentally that the best trade-off between these competing aspects is achieved by optimizing the demerit, which aims to place rows and columns with similar cluster membership next to each other.

\section*{Acknowledgements}
 This research is supported by the EC H2020 RIA project SoBigData++ (871042) and the Wallenberg AI, Autonomous Systems and Software Program (WASP) funded by the Knut and Alice Wallenberg Foundation. Some of the computations were enabled by
the National Academic Infrastructure for
Supercomputing in Sweden (NAISS) 
and Swedish National Infrastructure for
Computing (SNIC)
partially funded by
the Swedish Research Council through grant agreements no.~2022-06725 and 2018-05973.
\clearpage

\section*{Ethical Statement}

Boolean matrix factorization and similar pattern mining techniques can be used
to identify tightly-knit communities (biclusters) from bipartite networks, which
can help authorities to identify terrorist networks or dissidents from
inter-communication networks. They can also be used to identify people's
political opinions (e.g., by studying their social network behavior). The
visualization methods discussed in this paper cannot be used for  these actions
directly, as they require the underlying mining algorithm, but they do
facilitate the use of the mining algorithms in benign as well as nefarious
purposes. We do consider that the overall positive effects of these methods
greatly outweigh the problems caused by the as-such unavoidable negative use
cases.

\clearpage

\bibliographystyle{splncs04}
\bibliography{main}

\clearpage

\clearpage
\appendix

\section{Pseudocode}
\label{sec:pseudocode}

\subsection{Pseudocode for the Greedy Algorithms}
\label{sec:pseudocode-greedy}

We present the pseudocode for the algorithms from Sect.~\ref{sec:greedy} in
Algorithms~\ref{alg:greedy} and~\ref{alg:greedyAddRow}.

In the pseudocode, we write $\abs{\sigma_R}$ and $\abs{\sigma_C}$ to denote the number of elements
which are currently contained in the permutation.  When we add elements to the
permutations, we write $\sigma_R[i:]+[b]+\sigma_R[:i]$ to denote
the permutation $\sigma_R$ with element $b$ added at the $i$th position. 

We use \textsc{partialScore} to denote the objective function values, but only defined on the elements which are contained in the (partial) permutations $\sigma_R$ and $\sigma_C$, i.e., we ignore elements that have not yet been added to the permutations. For instance, for the greedy algorithm optimizing $f_{\clusterArea}$, $\Call{partialScore} {\sigma_R,\sigma_C}$ will return the value of $f_{\clusterArea}$ restricted to $\sigma_R$ and $\sigma_C$.

We define \textsc{greedyAddRow} in Algorithm~\ref{alg:greedyAddRow}. \textsc{greedyAddColumn} is defined in a similar manner, by adding the element $b$ in the column permutation $\sigma_C$ instead of in the row permutation $\sigma_R$.

\begin{algorithm}[t]
\caption{Greedy algorithm}
\label{alg:greedy}
\begin{algorithmic}
\Require $\mathcal{B}^R, \mathcal{B}^C$
\State $\sigma_R \gets \emptyset$, $\sigma_C \gets \emptyset$
\State Sort $\mathcal{B}^R$ and $\mathcal{B}^C$ by importance score
\ForAll{$i\in \{0, \dots, \max(|\mathcal{B}^R|, |\mathcal{B}^C|)\}$}
\If {$|\sigma_R| < |\mathcal{B}^R|$}
\State $\sigma_R \gets \Call{greedyAddRow}{\sigma_R, \sigma_C, \mathcal{B}^R[i]}$
\EndIf

\If {$|\sigma_C| < |\mathcal{B}^C|$}
\State $\sigma_C \gets \Call{greedyAddColumn}{\sigma_R, \sigma_C, \mathcal{B}^C[i]}$
\EndIf
\EndFor
\State \Return $\sigma_R,\sigma_C$
\end{algorithmic}
\end{algorithm}

\begin{algorithm}[t]
\caption{greedyAddRow}
\label{alg:greedyAddRow}
\begin{algorithmic}
\Require $\sigma_R,\sigma_C,b$
\State $l^* \gets \Call{partialScore}{\sigma_R+[b],\sigma_C}$
\State $i^* \gets \abs{\sigma_R}$
\For {$i=0,\dots,\abs{\sigma_R}-1$}
\State $l\gets \Call{partialScore}{\sigma_R[i:]+[b]+\sigma_R[:i],\sigma_C}$
\If{$l > l^*$}
\State $i^* \gets i$
\State $l^* \gets l$
\EndIf
\EndFor
\State $\sigma_R \gets \sigma_R[i^*:]+[b]+\sigma_R[:i^*]$
\State \Return $\sigma_R$
\end{algorithmic}
\end{algorithm}

\subsection{Pseudocode for Demerit-Based Algorithms}
\label{sec:greedy-demerit}
Algorithm~\ref{alg:tsp} presents the details of our TSP heuristic for minimizing
the demerit. Algorithm~\ref{alg:greedy-demerit} presents the details of our
greedy algorithm for minimizing the demerit, which is inspired by the algorithm
of~\cite{colantonio2011visual}. We define $\Call{demerit}{\sigma_i,\sigma_j}$ based on the definition of $\demerit(.)$ in Section~\ref{sec:demerit}:
\begin{align*}
	\Call{demerit}{\sigma_i,\sigma_j}
	= \sum_{b^R \in \mathcal{B}^R}
		\demerit(b^R; \sigma_i, \sigma_j),
\end{align*}

\begin{algorithm}[t]
\caption{TSP heuristic}
\label{alg:tsp}
\begin{algorithmic}
\Require $\mathcal{B}^C = \{b_1^C,\dots,b_t^C\}$
\State Create a distance matrix~$M$ with $M_{i,j} = \sum_{b^R \in \mathcal{B}^R} \Call{demerit}{b^R; b_i^C,b_j^C}$
\State Run a TSP solver on the distance matrix~$M$ to obtain a TSP tour~$(b_{i_1}^C,\dots,b_{i_t}^C)$
\State Find the best index~$j$ that optimizes the cluster area (Equation~\eqref{eq:cluster-area})
\State \Return $\sigma_C = (b_{i_j}^C, b_{i_{j}+1}^C, \dots, b_{i_{j}-1}^C)$
\end{algorithmic}
\end{algorithm}

\begin{algorithm}[t]
\caption{Greedy demerit algorithm}
\label{alg:greedy-demerit}
\begin{algorithmic}
\Require $\mathcal{B}$
\State $\sigma \gets \emptyset$
\State Sort $\mathcal{B}$ by importance score
\ForAll{$b\in \mathcal{B}$}
\If {$\abs{\sigma} < 2$}
\State $\sigma \gets \sigma + [b]$
\Else
\If{$\Call{demerit}{b,\sigma[0]} < \Call{demerit}{b,\sigma[\abs{\sigma}-1]}$ }
\State $p \gets1$
\State $l \gets\Call{demerit}{b,\sigma[0]}$
\Else
\State $p \gets|\sigma|$
\State $l \gets\Call{demerit}{b,\sigma[\abs{\sigma}-1]}$
\EndIf
\For {$i=2,...,|\sigma|-1$}
\State $l_{prec} \gets\Call{demerit}{b, \sigma[i-1]}$
\State $l_{succ} \gets\Call{demerit}{b,\sigma[i]}$
\State $l_{curr} \gets\Call{demerit}{\sigma[i-1],\sigma[i]}$
\If{$\min(l_{prec}, l_{succ})>l$ and $\max(l_{prec},l_{succ}) \leq l_{curr}$}
\State $p \gets i$
\State $l \gets\min(l_{prec}, l_{succ}$)
\EndIf
\EndFor
\State $\sigma \gets\sigma[p:] + [b] + \sigma[:p]$
\EndIf
\EndFor
\State \Return $\sigma$
\end{algorithmic}
\end{algorithm}

\end{document}